\documentclass{article}

\usepackage[preprint]{neurips_2025}

\usepackage[utf8]{inputenc} 
\usepackage[T1]{fontenc}    
\usepackage{hyperref}       
\usepackage{url}            
\usepackage{booktabs}       
\usepackage{amsfonts}       
\usepackage{nicefrac}       
\usepackage{microtype}      
\usepackage{xcolor}         

\usepackage{graphicx}
\usepackage{amsmath}
\usepackage{subcaption}

\title{M3: High-fidelity Text-to-Image Generation via Multi-Modal, Multi-Agent and Multi-Round Visual Reasoning}

\author{%
  \textbf{Bangji Yang}\thanks{Equal contribution.} , 
  \textbf{Ruihan Guo}, 
  \textbf{Jiajun Fan}, 
  \textbf{Chaoran Cheng}, 
  \textbf{Ge Liu}\thanks{Corresponding author.} \\
  University of Illinois Urbana-Champaign \\
  \texttt{\{bangjiy2, ruihang6, jiajunf3, chaoran7, geliu\}@illinois.edu} \\
}

\begin{document}

\maketitle

\begin{abstract}
  Generative models have achieved impressive fidelity in text-to-image synthesis, yet struggle with complex compositional prompts involving multiple constraints. We introduce \textbf{M3 (Multi-Modal, Multi-Agent, Multi-Round)}, a training-free framework that systematically resolves these failures through iterative inference-time refinement. M3 orchestrates off-the-shelf foundation models in a robust multi-agent loop: a Planner decomposes prompts into verifiable checklists, while specialized Checker, Refiner, and Editor agents surgically correct constraints one at a time, with a Verifier ensuring monotonic improvement. Applied to open-source models, M3 achieves remarkable results on the challenging OneIG-EN benchmark, with our Qwen-Image+M3 surpassing commercial flagship systems including Imagen4 (0.515) and Seedream 3.0 (0.530), reaching state-of-the-art performance (0.532 overall). This demonstrates that intelligent multi-agent reasoning can elevate open-source models beyond proprietary alternatives. M3 also substantially improves GenEval compositional metrics, effectively doubling spatial reasoning performance on hardened test sets. As a plug-and-play module compatible with any pre-trained T2I model, M3 establishes a new paradigm for compositional generation without costly retraining.
\end{abstract}

\begin{figure*}[ht]
  \centering
   \includegraphics[width=\linewidth]{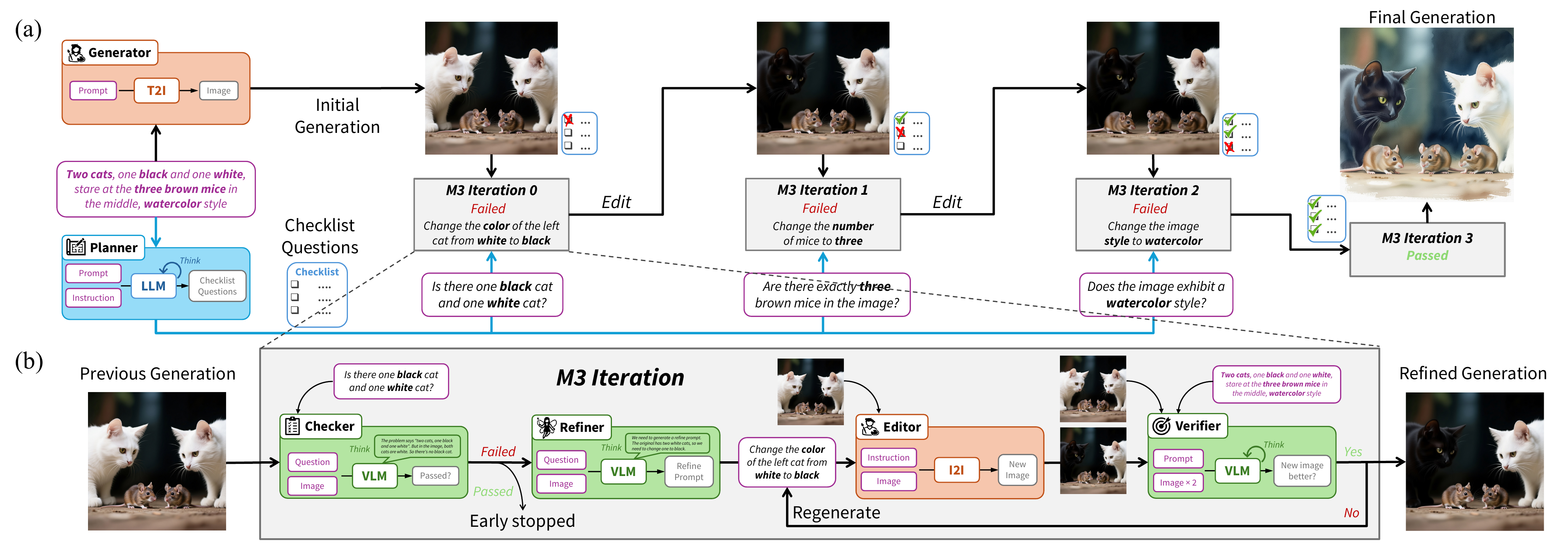}
   \caption{The General Framework of \textbf{M3 (Multi-Round, Multi-Agent, Multi-Modal)}: An Agentic Pipeline for Inference-Time Self-Refinement.
\textbf{(a)} The Multi-Round Optimization Process (Macro-View): Illustrates the high-level, progressive enhancement workflow. A LLM-based \textbf{Planner} agent first analyzes the complex user prompt to generate a \emph{checklist} of verifiable constraints. This checklist then orchestrates a \emph{multi-round} series of iterative edits, where each round targets a specific, detected alignment failure (e.g., Iteration 0 corrects \emph{attribute binding}, Iteration 1 corrects \emph{object count}, Iteration 2 corrects \emph{artistic style}), progressively resolving errors to produce the final refined generation.
\textbf{(b)} The Multi-Agent Workflow within a Single Round (Micro-View): Details the inner mechanics of any single M3 Iteration, revealing the \textbf{Multi-Agent} closed-loop feedback system. This pipeline consists of four collaborating agents that execute the plan from (a): 1) The \textbf{Checker} (VLM) evaluates the image against one constraint. 2) If ``Failed,'' the \textbf{Refiner} (VLM) generates a targeted edit instruction. 3) The \textbf{Editor} (an off-the-shelf model) executes the edit to create a new candidate. 4) The \textbf{Verifier} (VLM) performs quality assurance, accepting the edit only if it measurably improves alignment over the previous generation, thus ensuring \emph{monotonic enhancement}.}
\label{fig:General Framework}
\end{figure*}

\section{Introduction}
\label{sec:intro}

Text-to-Image (T2I) generative models, including diffusion-based systems, have achieved unprecedented success in synthesizing high-fidelity yet creative images \cite{rombach2022high, esser2024sd35large, podell2023sdxl, openai2023dalle3, qwen_image}. Despite their general success, a critical gap remains: these models frequently fail to adhere to complex user prompts, particularly those involving essential attributes (compositional elements, color, shape, precise object counts, and attribute binding), specific spatial relationships, text rendering, and knowledge reasoning \cite{chang2025oneig, wang2025mint}. This ``prompt-alignment failure'' results in generated images that omit, swap, or incorrectly place elements, failing to meet critical user constraints.

Current attempts to solve this alignment problem suffer from significant limitations. Open-loop prompt enhancement \cite{wu2025reprompt} cannot verify or correct visual output. Monolithic model retraining \cite{chen2024ipr} is inflexible and cannot be applied to the vast ecosystem of existing pre-trained models. Finally, iterative re-generation frameworks \cite{yang2024idea2img} often employ inefficient trial-and-error, discarding flawed candidates rather than progressively fixing them.

To address these limitations and unleash the power of current T2I models, we introduce \textbf{M3} (\textbf{Multi-Modal, Multi-Agent, Multi-Round}), a new paradigm for high-fidelity text-to-image generation that operates via \textbf{inference-time self-refinement}. M3 introduces a training-free, closed-loop agentic framework that progressively refines a generated image to match a complex prompt. As a plug-and-play module, it orchestrates a collaboration between off-the-shelf foundation models (VLMs and image editors) and tools, eliminating the need for costly retraining.

The core of M3 is its multi-agent reasoning pipeline, which systematically decomposes the complex alignment task into five well-defined procedures: 1) a \textbf{Checklist Planner} analyzes the prompt to identify key constraints and propose a verification checklist; 2) a \textbf{Checker} evaluates the candidate image against these questions; 3) a \textbf{Refiner} generates a concrete, actionable edit prompt upon detecting a misalignment; 4) an \textbf{Editor} refines the previous image based on this new instruction; and 5) a \textbf{Verifier} compares the edited image against the previous one, ensuring progressive enhancement and preventing quality degradation.

This multi-round, closed-loop approach is fundamentally more precise and efficient than standard regeneration. Instead of discarding flawed outputs, M3's planned refinement process effectively addresses complex alignment failures (such as attribute binding, spatial relations, and object counts) step by step, ensuring the final image remains faithful to the user's most detailed constraints. Our contributions are fourfold and can be summarized as follows:
\begin{enumerate}
    \item We introduce \textbf{M3}, a novel, training-free, and closed-loop agentic framework that introduces an \textbf{inference-time self-refinement} paradigm. Its core is a robust multi-agent reasoning pipeline---comprising a Planner, Checker, Refiner, Editor, and Verifier---that systematically decomposes complex prompts, identifies specific errors, and performs progressive, validated edits to ensure monotonic improvement.
    
    \item We demonstrate the framework's \textbf{universal applicability and model-agnosticism}, proving its effectiveness across different model scales and capabilities. We package M3 as \textbf{AutoRefiner}, a lightweight, plug-and-play module to enhance any T2I model without retraining. We validate this by enhancing both massive, state-of-the-art generators (e.g., Qwen-Image-20B) with a fully VLM-driven pipeline and smaller, specialized models (e.g., MindOmni-7B) with a lightweight, tool-augmented pipeline (\textbf{M3-Hybrid}).
    
    \item We achieve \textbf{state-of-the-art performance} through extensive experiments on multiple challenging compositional benchmarks (GenEval \cite{ghosh2023geneval} and OneIG-EN \cite{oneig2025}). Our results show \textbf{dramatic and comprehensive improvements} in fidelity, particularly on the most difficult tasks. Notably, M3 effectively \textbf{doubles the spatial reasoning performance} of a SOTA baseline on a hardened test set and achieves significant gains in the critical 'Alignment' and 'Reasoning' dimensions of OneIG-EN.
    
    \item Through extensive qualitative analysis, we demonstrate that M3 \textbf{systematically rectifies a wide range of common failure modes} that persist even in powerful SOTA models like Qwen-Image-20B. This includes correcting fundamental errors in spatial relationships, attribute binding, object counting, text rendering, style adherence, and complex multi-constraint composition.
\end{enumerate}

\section{Related Work}
\label{sec: related work}


\subsection{Text-to-Image Generative Models}
The dominant paradigm in T2I synthesis is the diffusion model~\cite{sohl2015deep,ho2020denoising}, particularly Latent Diffusion Models (LDM)~\cite{rombach2022high} which operate in a compressed latent space. This architecture forms the backbone of open-source models like Stable Diffusion~\cite{sd,podell2023sdxl,esser2024sd35large} and commercial systems. Recent large-scale models, such as DALL-E 3~\cite{openai2023dalle3}, Qwen-Image~\cite{qwen_image}, Imagen4~\cite{google_imagen_2025}, and Seedream 3.0~\cite{gao2025seedream}, have pushed the boundaries of photorealism and prompt-following, often by co-training with highly descriptive captions. Unified multimodal models like Show-o~\cite{xie2024show}, Show-o2~\cite{xie2025showo2}, Janus-Pro~\cite{chen2025janus}, GOT~\cite{fang2025got}, and OmniGen~\cite{xiao2025omnigen} integrate vision understanding and generation within a single transformer architecture. Despite these advances, all T2I models exhibit a persistent long tail of compositional failures, particularly in attribute binding, spatial relationships, and object counting, as measured by benchmarks like GenEval~\cite{ghosh2023geneval} and OneIG-EN~\cite{oneig2025}. Unlike approaches that propose new model architectures requiring extensive training, our work introduces a training-free, plug-and-play framework that systematically corrects compositional errors of any existing pre-trained T2I model through multi-round visual reasoning.



\subsection{Reasoning with Images}
The ability to reason about visual content is foundational to our framework. Modern Vision-Language Models (VLMs) pre-trained on large-scale image-text corpora~\cite{radford2021learning} have demonstrated remarkable capabilities in visual understanding. Models like Qwen-VL~\cite{qwenvl25,qwen3,qwen3omni} and MindOmni~\cite{xiao2025mindomni} can perform detailed image captioning, visual question answering (VQA), and semantic reasoning about complex visual scenes. While diagnostic tools like PUNC~\cite{franchi2025punc} leverage VLMs to quantify text-image misalignment through systematic questioning, they provide no mechanism to actively correct the detected errors, leaving users with diagnostic information but no solution. In contrast, M3 operationalizes VLM reasoning capabilities into a concrete, actionable multi-agent system where the VLM serves multiple coordinated roles—analyzing prompts as a Planner, detecting errors as a Checker, generating precise edit instructions as a Refiner, and validating improvements as a Verifier—forming a closed feedback loop for iterative correction rather than mere diagnosis.

\subsection{Self-Refinement in Generation}Several strategies have emerged to iteratively improve generative output, a concept broadly known as self-refinement.
\begin{itemize}

\item \textbf{Open-loop prompt enhancement} methods like RePrompt~\cite{wu2025reprompt} train LLMs to rewrite prompts but cannot detect or self-correct when enhanced prompts still produce errors. 

\item \textbf{Training-time correction} approaches like IPR~\cite{chen2024ipr} fine-tune models on corrected pairs, but their objective is to relabel prompts to match flawed images rather than fix images to match user intent, requiring costly retraining per base model. 

\item \textbf{Inference-time iteration} represents the most comparable category. Single-agent orchestrators like GenArtist employ a monolithic MLLM for planning, tool-calling, and verification—creating a complex single point of failure where all capabilities are entangled. 
Idea2Img~\cite{yang2024idea2img} uses multi-round VLM refinement, but its reflection is fed back to the prompt generator, which essentially does prompt optimization instead of reasoning at the image level. It is also inefficient to generate $N$ completely new prompts and images per round in order to ensure the improvement. 
Autoregressive methods~\cite{guo2025cot} explore token-level verification within the generation process itself, requiring model-internal access and limiting compatibility. Specialist frameworks like Anywhere~\cite{chen2024anywhere} use multi-agent loops but focus narrowly on mask-based inpainting for foreground-conditioned generation.
\end{itemize}
In contrast, our M3 establishes a unique paradigm through three key architectural innovations: (1) \textbf{Multi-agent modularity}—five specialized agents (Planner-Checker-Refiner-Editor-Verifier) with clear separation of concerns, enabling robustness and extensibility; (2) \textbf{Progressive surgical editing}—VLM-generated targeted edit instructions that fix specific detected errors rather than regenerating entire images, preserving correctly-generated elements; (3) \textbf{Closed-loop quality assurance}—a dedicated Verifier agent that compares each edit against the previous-best image, ensuring monotonic improvement and preventing the spurious modifications common in instruction-guided editing. This training-free, plug-and-play design systematically decomposes complex alignment tasks into verifiable steps, achieving superior efficiency and fidelity.

\section{Method}
\label{sec: method}

\begin{figure*}[ht]
  \centering
   \includegraphics[width=1.0\linewidth]{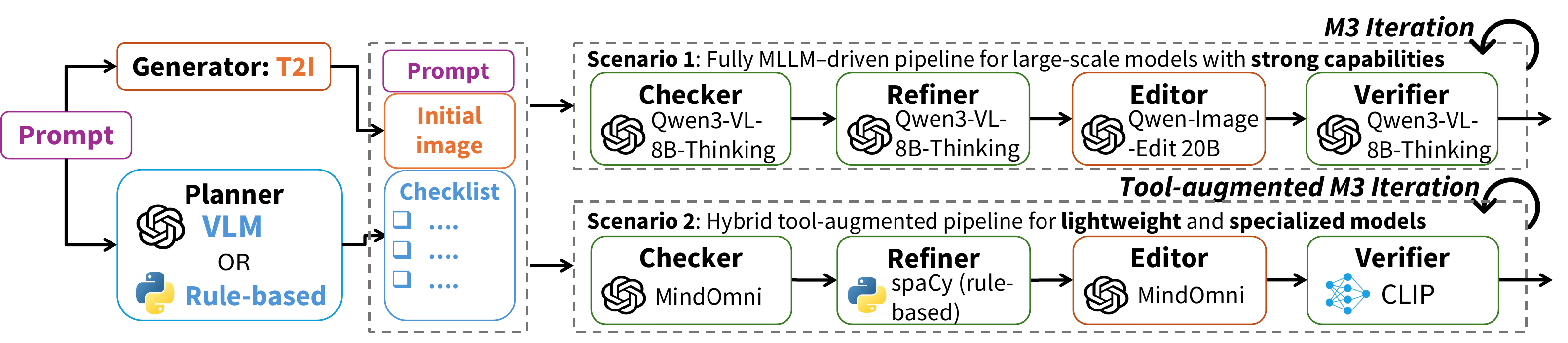}
   \vspace{-0.1in}
   \caption{M3 is a plug-and-play, model-agnostic framework that is compatible with both large-scale and lightweight VLMs and tools.}
   \label{fig:m3-versions}
\end{figure*}

We propose \textbf{M3}, a novel framework that decomposes the complex task of T2I-prompt alignment into a structured, multi-agent workflow. The M3 pipeline, depicted in Figure \ref{fig:General Framework}, is implemented as a training-free, closed-loop system that orchestrates the actions of two pre-trained, off-the-shelf foundation models: a Vision-Language Model (VLM) for all reasoning tasks and an instruction-guided Image Editor for visual manipulation.    

\subsection{The M3 Agentic Pipeline} 
As a training-free edit-based multi-agent framework, our M3 agentic pipeline starts with the initial prompt and image (iteration 0), which can be the output of any T2I model or a user-provided file.
The VLM's \emph{multi-agent} roles are then dynamically prompted as follows:

\textbf{Checklist Planner.} Given the user's initial prompt, the planner module generates a ``visual checklist'' of semantic constraints. With strong reasoning capabilities, the planner dynamically captures key attributes—color, quantity, shape, spatial relationships, text rendering, and knowledge-based reasoning. Questions are converted into a concise, numbered list serving as input to the Checker-Refiner.     

\textbf{Checker and Refiner.} The checker and refiner agents form the core reasoning loop in M3. For each checklist entry, the checker uses the current image as input, confirming alignment. The checker responds with ``PASS'' if the constraint is met.
If not met, the checker's response provides a concrete, actionable edit prompt to fix the inconsistency. This efficient, single-call design directly translates detected errors into solvable instructions.

\textbf{Editor.} The Editor agent is the image manipulation model (e.g., Qwen-Image-Edit \citep{qwen_image}) that receives the current image and editing instructions from the refiner, executing edits to produce a new candidate.

\textbf{Verifier.} To prevent quality degradation—where an edit might fix one error but harm overall coherence—M3 employs a Verifier agent. This VLM-based agent compares the previous-best image and new candidate according to the initial prompt.

This ``Multi-Round'' process, governed by the Verifier, ensures the final output is the product of progressive, validated enhancements.    

Our implementation utilizes \textbf{Qwen3-VL-8B-Thinking} \cite{qwenvl25} as the VLM, embodying all reasoning agents, and \textbf{Qwen-Image-Edit} \cite{qwen_image} as the Editor. See App. \ref{app: Experimental Details} for agent prompts.

\subsection{Multi-Round Iterative and Robust Refinement}

The ``Multi-Round'' design ensures the final output results from progressive, validated enhancements. Unlike single-pass pipelines, this iterative loop enforces monotonic improvement, preventing error accumulation or unintended degradations. It systematizes refinement by decoupling detection, correction, and validation into verifiable steps, making the process resilient to edit conflicts and model imperfections.   

Robustness is achieved through several mechanisms. First, the Verifier agent is critical: instead of accepting a local fix, it compares the new candidate against the previous-best image, ensuring the edit contributes to global quality and overall prompt alignment. Second, M3 prevents ``dead ends'' where the Editor may be unable to fix a specific constraint. If the Verifier rejects an edit, the framework attempts to regenerate multiple times (e.g., $K=2$ attempts). If all fail, an ``escape hatch'' activates, skipping that checklist item in future rounds to avoid infinite loops. Finally, VLM agents (Checker, Refiner) communicate using structured JSON format, ensuring machine-readable robustness (e.g., \{``passed'': false, ``reason'': ``Change the number of notes to 4, add an orange note''\}), preserving failure reasons and exact instructions for the Editor.

\subsection{M3-Hybrid: Tool-Augmented Extension for Lightweight VLMs}

To demonstrate M3's universality, we developed a hybrid variant for broad applicability, operating effectively with \textbf{lightweight, specialized} models rather than relying solely on large-scale VLMs. Our motivation is to show the core agentic logic—planning, checking, refining, and verifying—is model-agnostic and adaptable to various architectural constraints.

To achieve this with smaller models, we adopt a \textbf{tool-augmented} strategy, offloading specific roles to efficient, specialized modules. This allows the core lightweight model to focus on its strengths while external tools ensure iterative loop robustness.

We implement this using \textbf{MindOmni} \citep{xiao2025mindomni}, a unified multimodal architecture integrating visual understanding (ViT encoder) and generation (diffusion decoder). M3 roles are distributed as: 
\begin{itemize}
    \item \textbf{Planner and Refiner (Tool-Augmented):} We replace heavy reasoning VLM with a rule-based module using \textbf{spaCy} \citep{honnibal2020spacy}, efficiently parsing prompts to generate fixed checklists and template-based refinement instructions. 
    \item \textbf{Checker (MindOmni):} We leverage MindOmni's native \emph{Vision Understanding} interface to perform VQA tasks, evaluating whether the generated image meets checklist constraints.
    \item \textbf{Editor (MindOmni):} Using MindOmni's native \emph{Multimodal Generation} interface to refine images according to instructions.
    \item \textbf{Verifier (Tool-Augmented):} To maintain efficiency, we substitute VLM-based verifier with \textbf{CLIP} score \citep{ilharco2021openclip}, serving as a fast, quantitative proxy for measuring semantic alignment between prompt and new candidate image.
\end{itemize} 
This configuration demonstrates M3 pipeline improves alignment even without state-of-the-art reasoning models, proving the framework's strength lies in its structured agentic workflow, successfully coordinating diverse specialized components.

\subsection{AutoRefiner: A Plug-and-Play Package} 
We have packaged our primary M3 implementation (the Qwen-based pipeline) as \textbf{AutoRefiner}, a lightweight, easy-to-use Python library designed for maximal community utility.

The plug-and-play package is model-agnostic and flexibly integrated into any existing pipeline, working with images from any T2I model or user-provided files to enhance compositional fidelity. Once the ImageRefiner class is initialized, the entire multi-round agentic workflow—including quality checking, refinement, and validation—can be executed with \textbf{a single line of code}.

\section{Experiment}
\label{sec: experiment}

To validate M3's effectiveness, we conduct comprehensive experiments on two challenging compositional T2I benchmarks: \textbf{GenEval} \citep{ghosh2023geneval} and \textbf{OneIG-EN} \citep{oneig2025}. We demonstrate our framework's universal, plug-and-play capability by evaluating two primary configurations applied to different base models. First, our \textbf{M3-Hybrid} variant, using lightweight tools for efficiency, is applied to MindOmni. Second, our full \textbf{M3} framework is applied to SOTA Qwen-Image, testing both a \textbf{Rule-based Planner} (+M3 (Rule)) and full \textbf{VLM Planner} (+M3 (VLM)). Results demonstrate all M3 configurations significantly improve prompt-following fidelity and achieve state-of-the-art performance. See App. \ref{app: Experimental Details} for experimental details and App. \ref{app:additional_results} for more results.

\subsection{Benchmark Results on GenEval}
\label{sec:exp_geneval}

GenEval \citep{ghosh2023geneval} is a foundational benchmark designed to systematically evaluate \emph{compositional} capabilities of T2I models, addressing limitations of holistic metrics like FID or CLIPScore. It uses a multi-stage, object-focused pipeline to precisely measure adherence to specific compositional constraints, including \textit{object co-occurrence}, \textit{count}, \textit{spatial position}, and \textit{attribute binding}. The benchmark's high correlation with human judgment has established it as a standard tool for diagnosing fine-grained failure modes.

\subsubsection{Tool-Augmented M3: Enhancement for Specialized Models}
\label{sec:exp_mindomni}

To demonstrate M3's \textbf{universality} and \textbf{flexibility}, we first evaluate its tool-augmented variant, \textbf{M3-Hybrid}. This configuration is designed for computational efficiency, replacing heavyweight VLM agents with lightweight, specialized tools (e.g., spaCy for planning, CLIP for verification). We apply M3-Hybrid to the 7B-parameter MindOmni model \citep{xiao2025mindomni}, combining these tools with the model's native VQA and T2I capabilities while maintaining the core plan-check-refine-edit-verify logic.

We measured our MindOmni baseline using fixed inference parameters (50 steps, \texttt{max\_new\_tokens} 512, \texttt{guidance\_scale} 3.0). Results in Table~\ref{tab:geneval-mindomni} demonstrate M3-Hybrid's efficacy even with simplified components. The most striking improvement appears in the notoriously difficult \textbf{Attribute Binding} category, achieving \textbf{substantial gains}, while also clearly improving counting and position scores. This experiment establishes a crucial principle: \textbf{M3's architectural design is more important than its individual components}, proving framework robustness to successfully correct errors even on efficiency-constrained models.

\begin{table}[ht]
\centering
\caption{Results on \textbf{original GenEval}. M3 significantly enhances MindOmni's compositional performance. \textbf{Best} results are bold, \underline{second-best} underlined.}
\label{tab:geneval-mindomni}
\small
\resizebox{\linewidth}{!}{%
\begin{tabular}{@{}lcccccccc@{}}
\toprule
\textbf{Model} & \textbf{Size} & \textbf{Single} & \textbf{Two} & \textbf{color} & \textbf{count} & \textbf{pos} & \textbf{Attr} & \textbf{Ovr} \\
\midrule
SDv1-5 & 0.9B & 0.97 & 0.38 & 0.35 & 0.76 & 0.04 & 0.06 & 0.43 \\
SDXL & 2.6B & 0.98 & 0.74 & 0.39 & \underline{0.85} & 0.15 & 0.23 & 0.55 \\
OmniGen & 3.8B & \textbf{0.99} & 0.86 & 0.85 & 0.64 & 0.31 & 0.55 & 0.70 \\
Show-o & 1.5B & 0.95 & 0.52 & 0.49 & 0.82 & 0.11 & 0.28 & 0.53 \\
Janus-Pro & 7.0B & \textbf{0.99} & 0.89 & 0.59 & \textbf{0.90} & \textbf{0.79} & \underline{0.66} & \textbf{0.80} \\
GOT & 7.9B & \textbf{0.99} & 0.69 & 0.85 & 0.67 & 0.34 & 0.27 & 0.64 \\
\midrule
MindOmni (base) & 7.0B & \textbf{0.99} & \textbf{0.97} & \underline{0.90} & 0.59 & 0.57 & 0.60 & \underline{0.77} \\
\quad\textbf{+M3-Hybrid} & 7.0B & \textbf{0.99} & \textbf{0.97} & \textbf{0.91} & 0.64 & \underline{0.58} & \textbf{0.69} & \textbf{0.80} \\
\bottomrule
\end{tabular}
}
\end{table}

\subsubsection{Fully VLM-Powered M3: Maximum Fidelity for SOTA Models}
\label{sec:exp_qwen}

Having demonstrated M3's effectiveness with lightweight tools, we now evaluate its full agentic configuration on the 20B Qwen-Image model—a SOTA generator whose near-ceiling performance saturates the original GenEval benchmark. To address this evaluation gap, we created a \textbf{new, significantly harder GenEval test set} focused on complex compositional scenarios. This new benchmark decomposes the monolithic ``Attribute Binding'' category into three granular sub-tasks: \textit{color\_attr}, \textit{counting\_attr}, and the highly challenging \textit{color\_position\_attr}. We test two M3 variants: one with a rule-based Planner and one with a VLM Planner.

Results on this hardened benchmark (Table~\ref{tab:geneval-qwen}) reveal M3 delivers dramatic improvements where the SOTA baseline fails. Most notably, \textbf{spatial reasoning performance on ``position'' effectively doubles} with the VLM Planner, transforming a near-failing capability into a competent one. Both variants achieve significant gains on the most difficult multi-attribute tasks, revealing a trade-off: the Rule-based Planner achieves \textbf{state-of-the-art results} on structured tasks (\textit{counting\_attr}), while the VLM Planner is \textbf{demonstrably superior} for complex spatial reasoning (\textit{color\_position\_attr}). This confirms M3's value in enhancing even the most powerful models by surgically correcting specific, persistent failures.

\begin{table}[ht]
\centering
\caption{Results on \textbf{hardened GenEval}. M3 dramatically improves Qwen-Image on complex compositional tasks, effectively doubling spatial reasoning performance.}
\label{tab:geneval-qwen}
\small
\resizebox{\linewidth}{!}{%
\begin{tabular}{@{}lccccccc@{}}
\toprule
\textbf{Model} & \textbf{Single} & \textbf{Two} & \textbf{color} & \textbf{count} & \textbf{pos} & \multicolumn{2}{c}{\textbf{Attr Binding}} \\
\cmidrule(l){7-8}
 & \textbf{Obj} & \textbf{Obj} & & & & \textbf{color\_attr} & \textbf{Overall} \\
\midrule
Qwen-Image (base) & \textbf{1.00} & \textbf{0.98} & \textbf{0.99} & 0.81 & 0.26 & 0.83 & 0.81 \\
\quad+M3 (Rule) & \textbf{1.00} & \textbf{0.98} & \textbf{0.99} & \textbf{0.86} & 0.49 & \textbf{0.84} & \underline{0.86} \\
\quad+M3 (VLM) & \textbf{1.00} & \textbf{0.98} & \textbf{0.99} & \underline{0.85} & \textbf{0.54} & \textbf{0.84} & \textbf{0.87} \\
\bottomrule
\end{tabular}
}
\vspace{-0.5em}
\end{table}

\subsection{Surpassing Commercial Flagship Models on OneIG-EN}
\label{sec:exp_oneig}

To complement our analysis on object-centric composition, we further evaluate our Qwen-M3 series on the \textbf{OneIG-EN} benchmark \citep{oneig2025}. OneIG-EN is a comprehensive benchmark assessing text-to-image generation quality across five critical axes: \textit{Alignment} (overall prompt fidelity), \textit{Text} (typographic rendering), \textit{Reasoning} (logical and knowledge-based concepts), \textit{Style} (artistic instruction adherence), and \textit{Diversity} (output variation). This benchmark allows us to test whether iterative refinement enhances prompt fidelity without degrading aesthetic quality.

\begin{figure*}[!ht]
    \centering
    \includegraphics[width=1.0\linewidth]{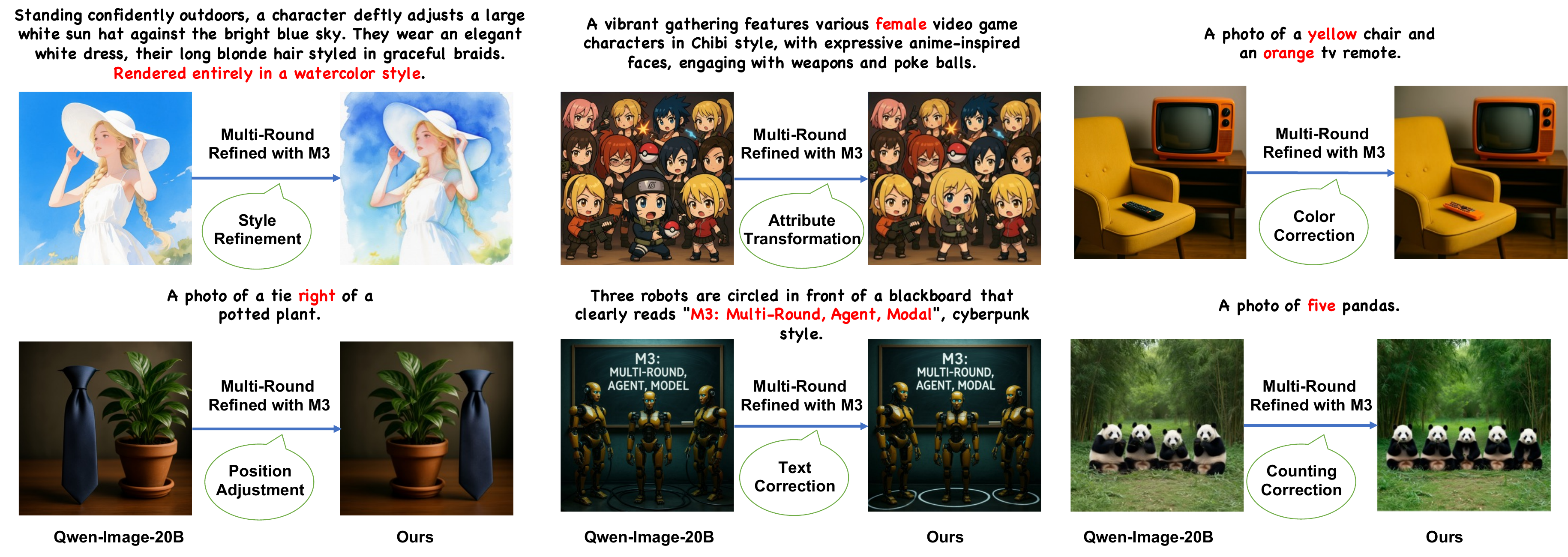}
    \caption{\textbf{M3 Systematically Corrects Diverse Failure Modes in State-of-the-Art Baseline Models.} Side-by-side comparison of Qwen-Image-20B and M3-refined outputs across six failure categories.}
    \label{fig:ours_vs_base}
\end{figure*}

\begin{figure*}[!ht]
    \centering
    \includegraphics[width=1.0\linewidth]{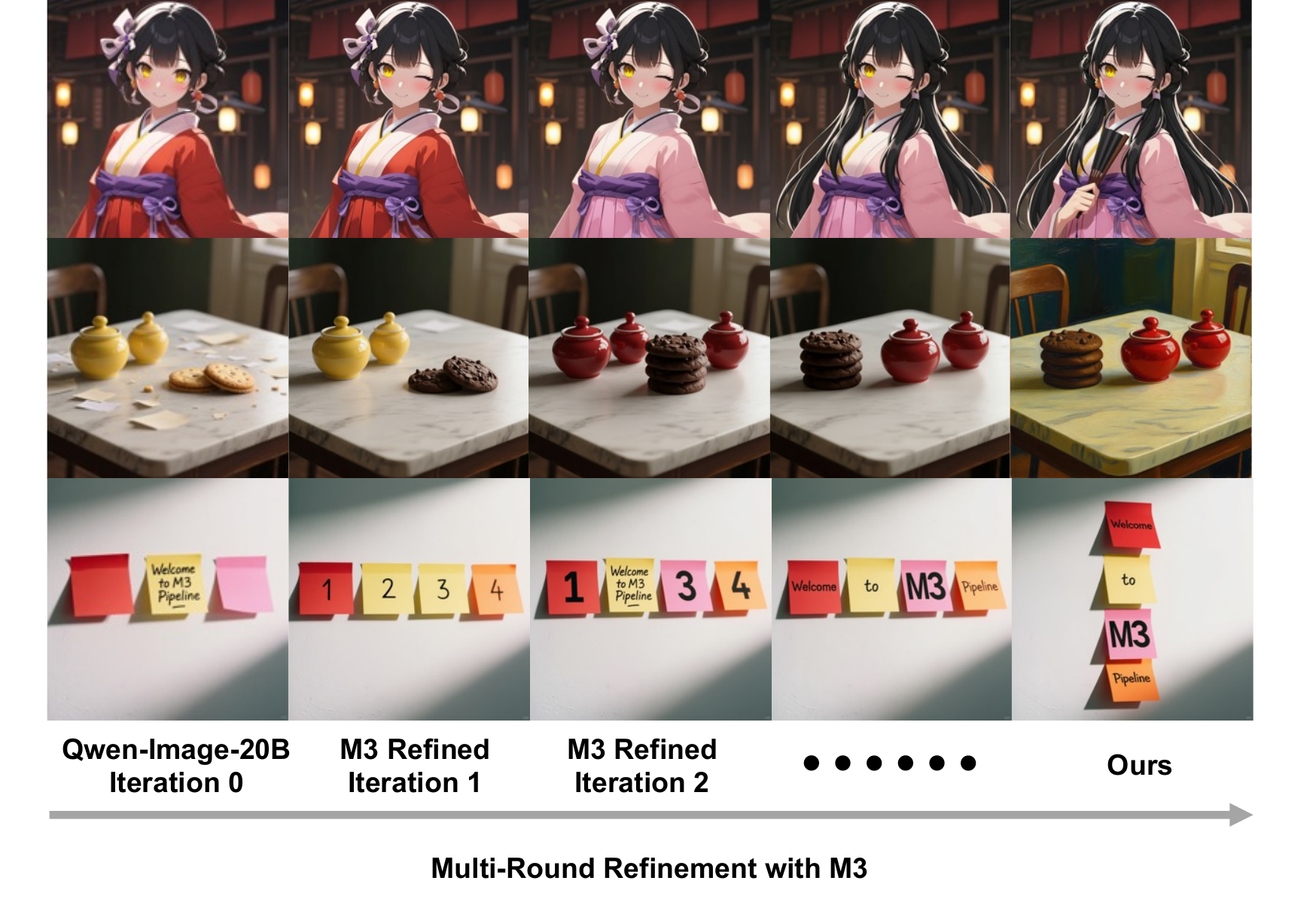}
    \caption{\textbf{Visualization of M3's Multi-Round Refinement Process.} Iterative enhancement from failed baseline (Iteration 0) to final aligned result for three highly complex compositional prompts.}
    \label{fig:multi_round}
\end{figure*}

\paragraph{Experimental Setup.}
We measured our Qwen-Image baseline using fixed inference parameters (40 steps, CFG scale 4.0). M3 variants are applied directly to this baseline's output, ensuring all observed improvements are attributable to M3 refinement.

\begin{table}[ht]
\centering
\caption{Results on OneIG-EN. M3 achieves state-of-the-art, \textbf{surpassing Imagen4 (0.515) and Seedream 3.0 (0.530)}.}
\label{tab:oneig}
\small
\resizebox{\linewidth}{!}{%
\begin{tabular}{@{}lcccccc@{}}
\toprule
\textbf{Model} & \textbf{Align.} & \textbf{Text} & \textbf{Reason.} & \textbf{Style} & \textbf{Div.} & \textbf{Overall} \\
\midrule
Show-o2-7B & 0.817 & 0.002 & 0.226 & 0.317 & 0.177 & 0.308 \\
SDXL & 0.688 & 0.029 & 0.237 & 0.332 & \textbf{0.296} & 0.316 \\
SD 3.5L & 0.809 & 0.629 & \underline{0.294} & 0.353 & 0.225 & 0.462 \\
Imagen4 & 0.857 & 0.805 & \textbf{0.338} & 0.377 & 0.199 & 0.515 \\
Seedream 3.0 & 0.818 & 0.865 & 0.275 & \underline{0.413} & \underline{0.277} & 0.530 \\
\midrule
Qwen-Image (base) & \textbf{0.885} & \underline{0.899} & 0.283 & 0.395 & 0.150 & 0.522 \\
\quad\textbf{+M3 (Rule)} & \underline{0.883} & 0.887 & 0.291 & \textbf{0.434} & 0.161 & \underline{0.531} \\
\quad\textbf{+M3 (VLM)} & 0.882 & \textbf{0.915} & 0.290 & 0.410 & 0.161 & \textbf{0.532} \\
\bottomrule
\end{tabular}
}
\vspace{-0.5em}
\end{table}

\paragraph{State-of-the-Art Performance.}
As shown in Table~\ref{tab:oneig}, both M3 variants achieve \textbf{state-of-the-art overall performance}, surpassing powerful commercial models including Seedream 3.0 (ByteDance, 0.530) and Imagen4 (DeepMind, 0.515). M3 (VLM) reaches 0.532 overall score, while M3 (Rule) achieves 0.531—both establishing new benchmarks for compositional generation. This is particularly remarkable given that our training-free, inference-time approach outperforms models requiring massive datasets and computational resources for training.

\paragraph{Targeted Improvements Across Key Dimensions.}
Comparing against the strong Qwen-Image baseline reveals M3's systematic enhancements. M3 (VLM) achieves the highest \textbf{Text rendering score across all models} (0.915), addressing one of the most challenging capabilities in generative AI. M3 (Rule) establishes \textbf{state-of-the-art Style adherence} (0.434), demonstrating that structured planning excels at capturing explicit stylistic constraints. Both variants show improvements in \textbf{Reasoning} and \textbf{Diversity}, while maintaining highly competitive Alignment scores. Notably, these gains come without aesthetic penalties—surgical refinement preserves visual quality while enhancing prompt adherence.

\paragraph{Complementary Planner Capabilities.}
The Rule and VLM Planners exhibit distinct strengths: Rule-based planning achieves superior style adherence through structured constraint checking, while VLM-based planning excels at complex text rendering and overall compositional reasoning. This validates M3's modular design, allowing users to select the optimal configuration for their specific generation requirements.

\subsection{Qualitative Analysis}
\label{sec:qualitative}

\subsubsection{Rectifying Failure Modes of Existing Models}

While quantitative results in Tables~\ref{tab:geneval-qwen} and \ref{tab:oneig} demonstrate M3's superior performance, Figure~\ref{fig:ours_vs_base} provides compelling visual evidence, showing M3 rectifying six systematic failure modes that persist even in the 20B-parameter Qwen-Image-20B. For \textit{``Standing confidently outdoors...Rendered entirely in a watercolor style''}, the baseline ignores the style instruction, producing a photorealistic portrait; M3 transforms it into an authentic watercolor painting while preserving semantic content. For \textit{``A vibrant gathering features various female video game characters''}, the baseline includes male characters; M3 performs precise transformations to achieve an all-female cast. For \textit{``A photo of a yellow chair and an orange tv remote''}, the baseline causes color-object inversion (orange chair, yellow remote); M3 resolves both binding errors while preserving the retro aesthetic.

The spatial prompt \textit{``A photo of a tie right of a potted plant''} results in the baseline placing the tie to the \textit{left}; M3 corrects this positional error while maintaining photorealism. For text rendering, \textit{``Three robots...clearly reads `M3: Multi-Round, Agent, Modal', cyberpunk style''} produces illegible baseline text; M3 generates the exact requested phrase clearly. Finally, \textit{``A photo of five pandas''} yields only four pandas in the baseline; M3 adds the missing fifth panda naturally. These examples span major compositional failure categories—style adherence, attribute binding, color correction, spatial reasoning, text rendering, and counting—proving M3's effectiveness as a training-free solution that surgically corrects diverse errors without degrading visual quality.

\begin{figure*}[ht]
    \centering
    \includegraphics[width=1.0\linewidth]{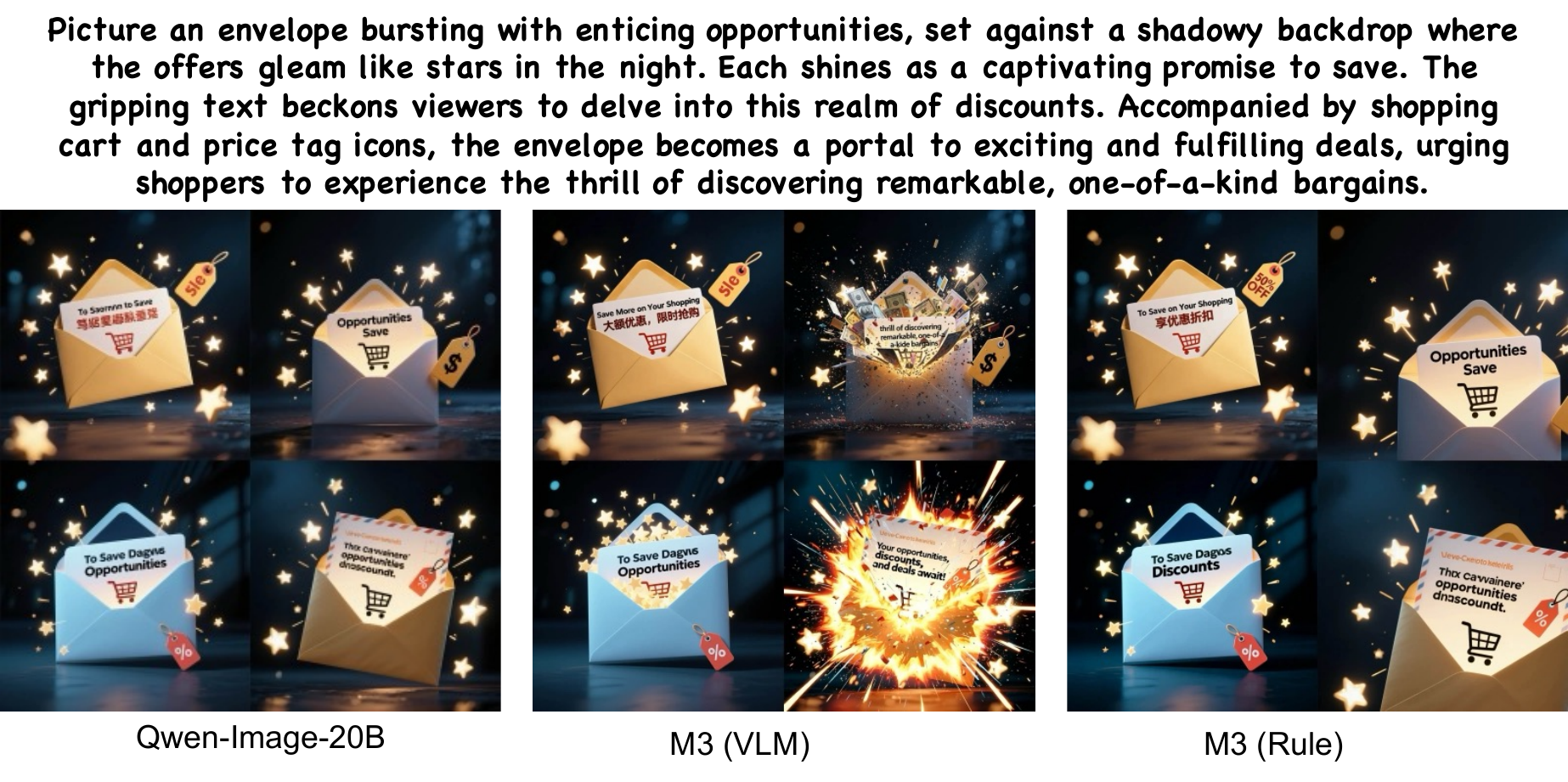}
    \caption{\textbf{Comparative Analysis of M3 Design Variants.} Visual comparison showing complementary strengths: M3 (VLM) excels at creative interpretation and complex text rendering (note dramatic visual effects), while M3 (Rule) maintains tighter compositional control with cleaner, structured layouts.}
    \label{fig:ab_diff_design}
\end{figure*}

\subsubsection{Visualizing the Multi-Round Refinement Process}
\label{sec:exp_multi_round}

While Figure~\ref{fig:ours_vs_base} demonstrates that M3 successfully rectifies baseline failures, Figure~\ref{fig:multi_round} provides deeper insight into \emph{how} it achieves this: by decomposing intractable, complex-conjunction prompts into a series of solvable, single-step edits. This multi-round process is key to handling high compositional complexity where single-pass generators fail.

\textbf{Attribute Conjunction.} The top row illustrates a prompt with extremely high attribute density (anime girl with specific eye color, hair style, and layered clothing). The baseline (Iteration 0) produces a generic image missing key attributes such as the \textit{``pink japanese hakama skirt''}. M3's iterations systematically stack corrections: Iteration 2 successfully corrects the skirt color to pink, with later iterations refining hair style (long hair with low twintails) and accessories to build towards the final, highly-aligned result.

\textbf{Spatial, Count, \& State.} The middle row demonstrates M3's ability to decouple and sequentially solve complex commands requiring \textit{``4 stacked chocolate cookies''} on the \textit{``left of 2 red porcelain''}. The baseline fails on multiple dimensions: cookies are unstacked and scattered, porcelain is yellow instead of red, and spatial arrangement is incorrect. M3's iterations exhibit surgical precision: Iteration 1 stacks the cookies correctly, Iteration 2 corrects the porcelain color to red, and the final image achieves proper spatial positioning (cookies on the left)—demonstrating monotonic enhancement where each correction preserves previous fixes.

\textbf{Joint Spatial \& Text Rendering.} The bottom row tackles joint spatial relations and multi-object text rendering, requiring four color-specific notes (\textit{``red, yellow, orange, and pink''}) with distinct words (\textit{``Welcome'', ``to'', ``M3'', ``Pipeline''}) arranged at \emph{``varying heights''}. The baseline produces only three notes with garbled text in incorrect arrangement. M3's iterations progressively correct count, color, text content, and spatial layout: the final image achieves the correct count (four notes), correct colors, correct spatial arrangement (varying heights), and legible text properly bound to each note.

\subsection{Ablation Results: Exploring Different Design Choices of M3}

To understand M3's design decisions, we ablate two key factors: planner implementation and iteration rounds.

\textbf{Planner Design: Rule-based vs. VLM-based.}
Table~\ref{tab:oneig} and Figure~\ref{fig:ab_diff_design} reveal complementary strengths. The \textbf{Rule-based Planner} employs deterministic, template-driven checklist generation, achieving superior \textit{Style} adherence (0.434 vs. 0.410) through systematic constraint matching. Figure~\ref{fig:ab_diff_design} confirms tighter compositional control and more faithful reproduction of discrete elements. The \textbf{VLM-based Planner} leverages dynamic reasoning for context-sensitive planning, achieving state-of-the-art \textit{Text} rendering (0.915 vs. 0.887) and superior spatial reasoning (position: 0.54 vs. 0.49, Table~\ref{tab:geneval-qwen}). As Figure~\ref{fig:ab_diff_design} shows, it produces more creative interpretations that capture implicit artistic intent beyond literal specifications.

\textbf{Number of Iterations.}
Figure~\ref{fig:multi_round} demonstrates monotonic improvement across iterations. Performance typically saturates at 3-5 iterations for most prompts, with the Verifier automatically stopping when no further gains are possible. For extremely complex prompts requiring multi-dimensional constraint resolution, additional iterations continue yielding measurable improvements as each round addresses one constraint dimension.

\section{Conclusion}
\label{sec:conclusion}

We propose M3, a training-free, multi-agent framework that systematically addresses compositional failures in text-to-image generation through closed-loop visual reasoning. Unlike approaches requiring costly retraining or wastefully discarding flawed outputs, M3 introduces surgical refinement via a structured plan-check-refine-edit-verify loop, orchestrating VLMs and image editors as specialized agents that progressively correct specific errors while preserving correctly-generated elements. M3 demonstrates universal applicability across model scales—from lightweight specialized models using tool-augmented components to massive state-of-the-art generators with full VLM reasoning. Comprehensive experiments validate M3's effectiveness: substantial improvements in attribute binding and counting on GenEval, effectively doubled spatial reasoning performance on hardened test sets, and state-of-the-art results on OneIG-EN surpassing powerful commercial models including Imagen4 and Seedream 3.0. As a plug-and-play module requiring no retraining, M3 establishes a new paradigm proving that intelligent, multi-agent refinement unlocks capabilities beyond single-pass generation.

{\small
\bibliographystyle{plainnat}
\bibliography{references}
}

\clearpage


\appendix


\appendix
\onecolumn

\section{Experimental Details}
\label{app: Experimental Details}

This section provides implementation details for reproducibility.

\subsection{Detailed Agent Prompts}

We provide the complete prompts used for each agent in the M3 pipeline:

\textbf{Planner Agent:} Given the user prompt $P$, generate a numbered checklist of verifiable visual constraints. Each item should be a yes/no question about a specific compositional element (object, attribute, spatial relation, count, style, text content).

\textbf{Checker Agent:} Given image $I$ and constraint $c_i$, determine if $c_i$ is satisfied. Respond with JSON: \{``passed'': true/false, ``reason'': ``explanation''\}. If false, the reason should be an actionable edit instruction.

\textbf{Refiner Agent:} (Implicit in Checker response when check fails)

\textbf{Editor Agent:} Execute the edit instruction on image $I$ to produce $I'$.

\textbf{Verifier Agent:} Given original prompt $P$, previous-best image $I_{\text{best}}$, and candidate $I'$, determine which better satisfies $P$. Respond with: ``better'', ``worse'', or ``same''.

\subsection{Hyperparameters}

For Qwen-Image baseline: 40 inference steps, CFG scale 4.0.
For MindOmni baseline: 50 inference steps, max\_new\_tokens 512, guidance\_scale 3.0.
For M3 refinement: Maximum 5 iterations, $K=2$ retry attempts per failed constraint.

\clearpage

\section{Algorithm Pseudocode}
\label{app: pseudocode}

\begin{quote}
\small
\textbf{Algorithm: M3 Multi-Round Refinement}
\begin{verbatim}
Input: prompt P, initial image I_0, VLM, Editor
Output: refined image I_final

1. checklist C = Planner(P)
2. I_best = I_0
3. for round = 1 to max_rounds:
4.   for each constraint c in C:
5.     if c already passed: continue
6.     check_result = Checker(I_best, c)
7.     if check_result.passed:
8.       mark c as passed
9.       continue
10.    edit_instr = check_result.reason
11.    for attempt = 1 to K:
12.      I_candidate = Editor(I_best, edit_instr)
13.      verify_result = Verifier(P, I_best, I_candidate)
14.      if verify_result == "better":
15.        I_best = I_candidate
16.        break
17.    if all attempts failed:
18.      mark c as skipped (escape hatch)
19.  if all constraints passed or skipped:
20.    break
21. return I_best
\end{verbatim}
\end{quote}

\clearpage

\section{Additional Experimental Results}
\label{app:additional_results}

In this section, we provide qualitative examples to supplement our quantitative findings.   

\subsection{Nuanced Alignment Beyond Automated Metrics}
As mentioned in the main text, there are instances where the alignment score remains unchanged, yet human evaluation confirms significant visual improvements. This discrepancy often arises because automated metrics may lack the granularity to perceive specific attribute modifications or structural clarifications.

\begin{figure}[h]
    \centering
    \begin{subfigure}{\linewidth}
        \centering
        \includegraphics[width=0.495\linewidth]{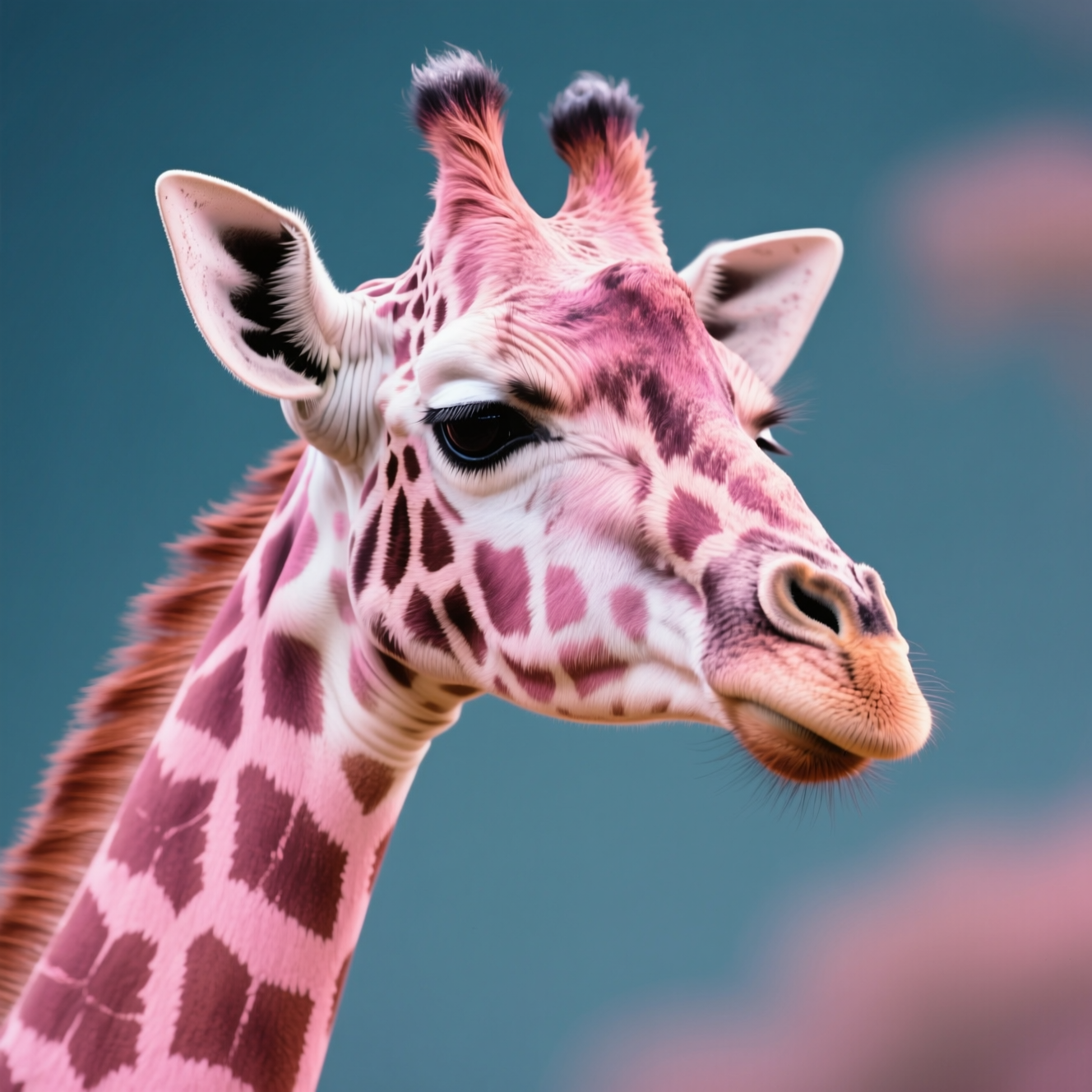}
        \hfill
        \includegraphics[width=0.495\linewidth]{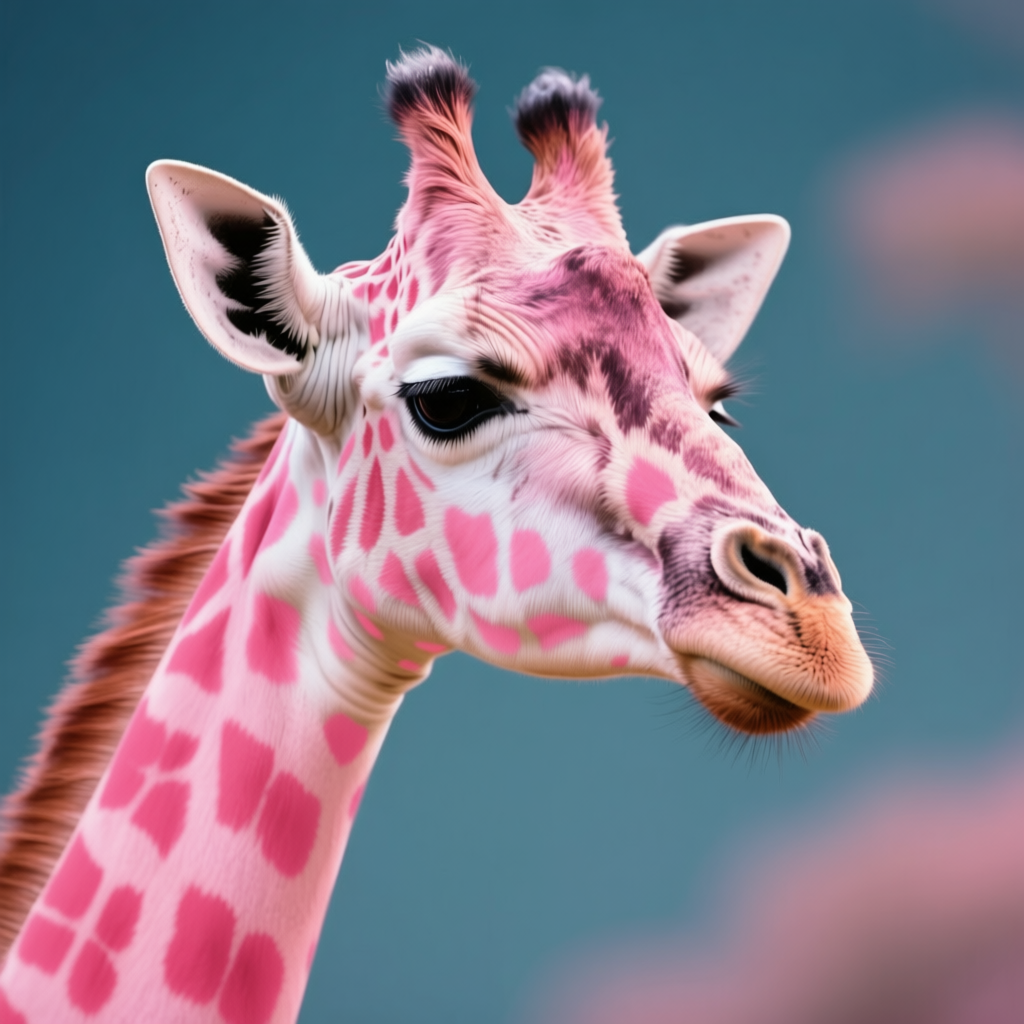}
        \caption{\textbf{Attribute Specificity:} The baseline (left) retains brown spots. M3 (right) explicitly executes the instruction: \textit{``Change the color of the giraffe's spots from brown to pink."}}
        \label{fig:giraffe}
    \end{subfigure}
    
    \vspace{1em}
    
    \begin{subfigure}{\linewidth}
        \centering
        \includegraphics[width=0.495\linewidth]{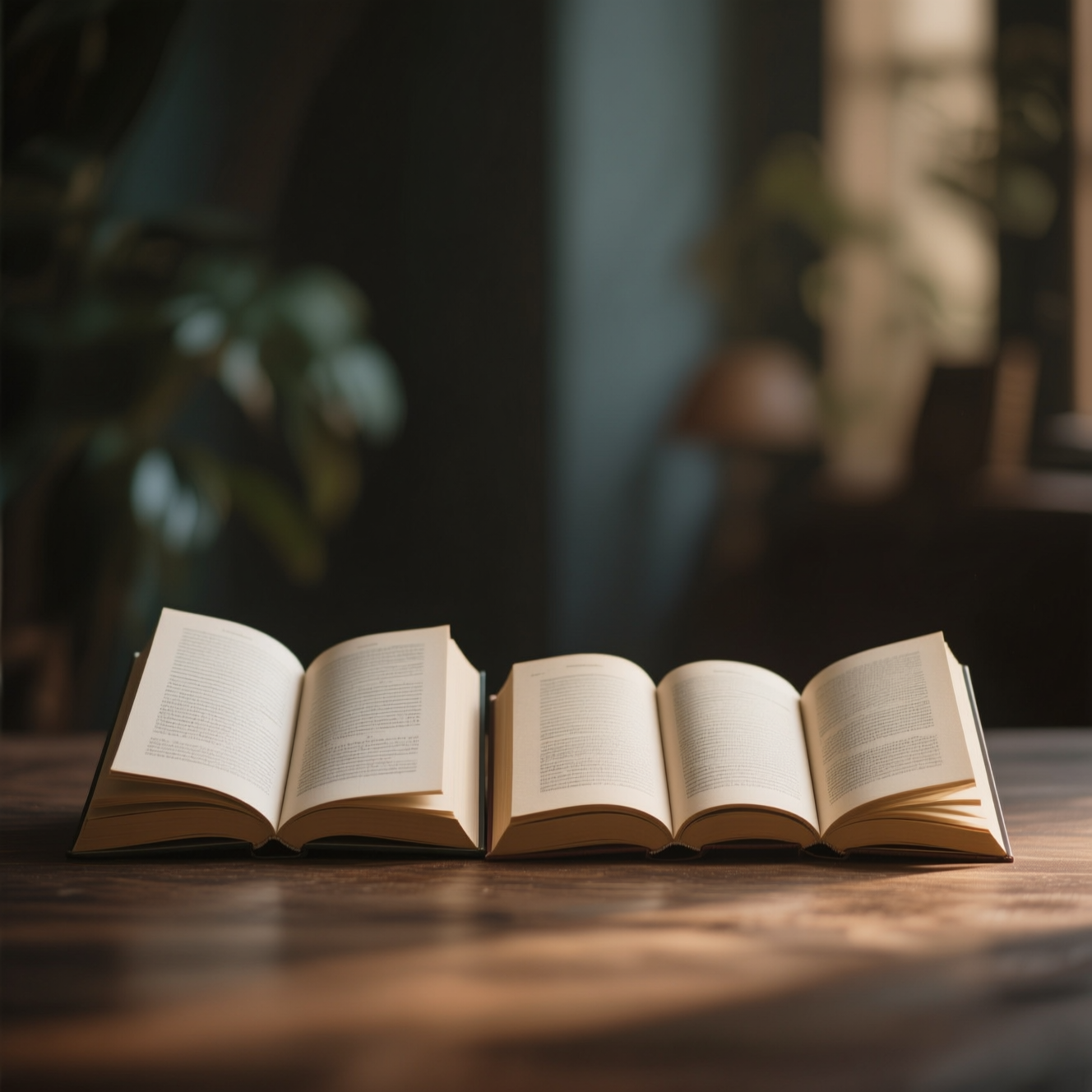}
        \hfill
        \includegraphics[width=0.495\linewidth]{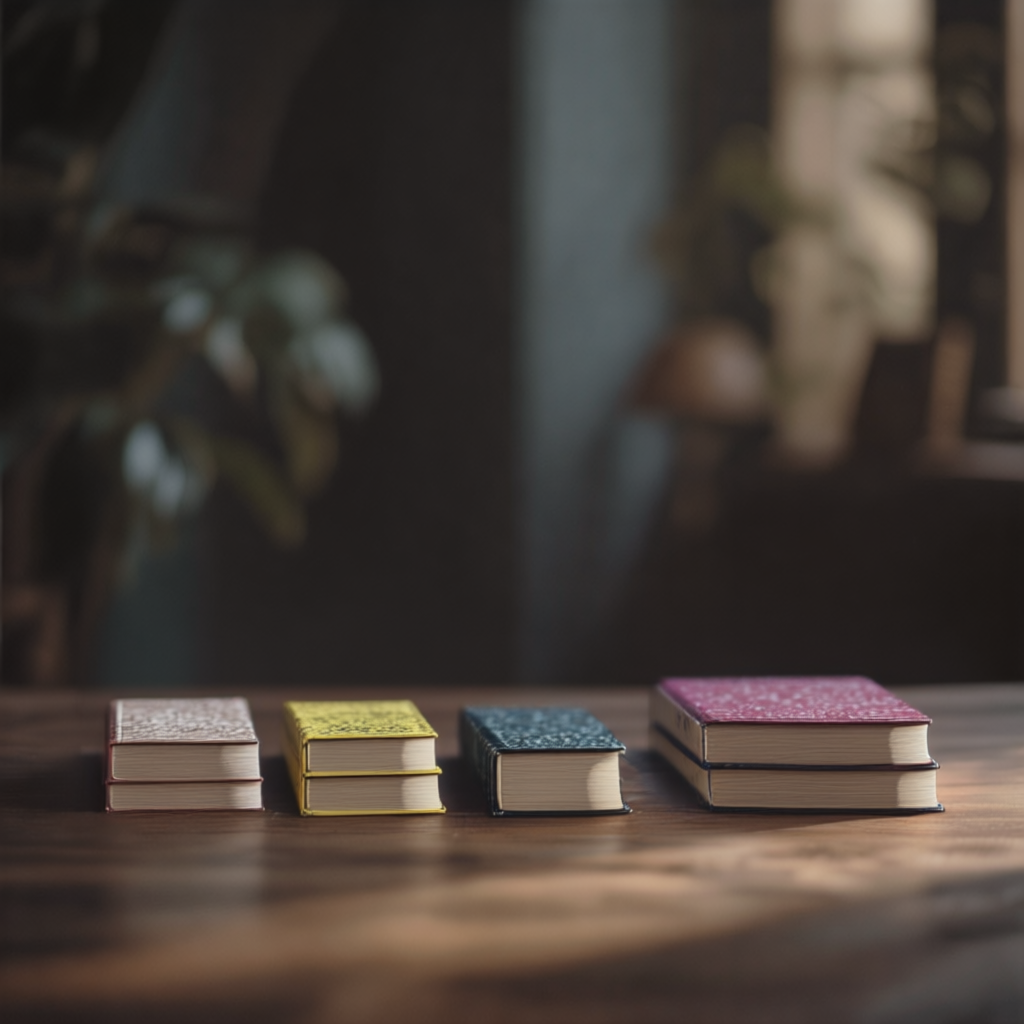}
        \caption{\textbf{Visual Clarity:} While the baseline (left) presents an ambiguous stack, M3 (right) refines the structure to make the count of ``four books" visually distinct.}
        \label{fig:books}
    \end{subfigure}
    \caption{\textbf{Nuanced Alignment.} Qualitative examples showing M3's ability to correct fine-grained attributes and structure.}
    \label{fig:qual_nuance}
\end{figure}

\noindent\textbf{Case 1: Attribute Specificity (The Pink Giraffe).} 
As shown in Figure \ref{fig:qual_nuance}(a), for the prompt ``\textit{a photo of a pink giraffe}," the baseline model generated a giraffe with a pinkish tint but retained natural brown spots. While this might pass a coarse check, it fails the user's specific intent. M3 successfully identified this attribute mismatch and generated a precise refinement instruction: \textit{``Change the color of the giraffe's spots from brown to pink,"} resulting in a much more faithful generation.

\noindent\textbf{Case 2: Visual Clarity (Four Books).}
In Figure \ref{fig:qual_nuance}(b), the prompt ``\textit{a photo of four books}" initially produced a cluttered image where individual books were indistinguishable. Although the refined result might still challenge a rigid object detector (leading to a similar quantitative score), M3 significantly improved the visual distinctness of the objects, making the ``four books" clearly identifiable to a human observer.

\subsection{Aesthetic Awareness and Logical Reasoning}
Beyond explicit instruction following, M3 demonstrates the ability to infer implicit constraints related to aesthetics and physical common sense.

\begin{figure}[h]
    \centering
    \begin{subfigure}{\linewidth}
        \centering
        \includegraphics[width=0.495\linewidth]{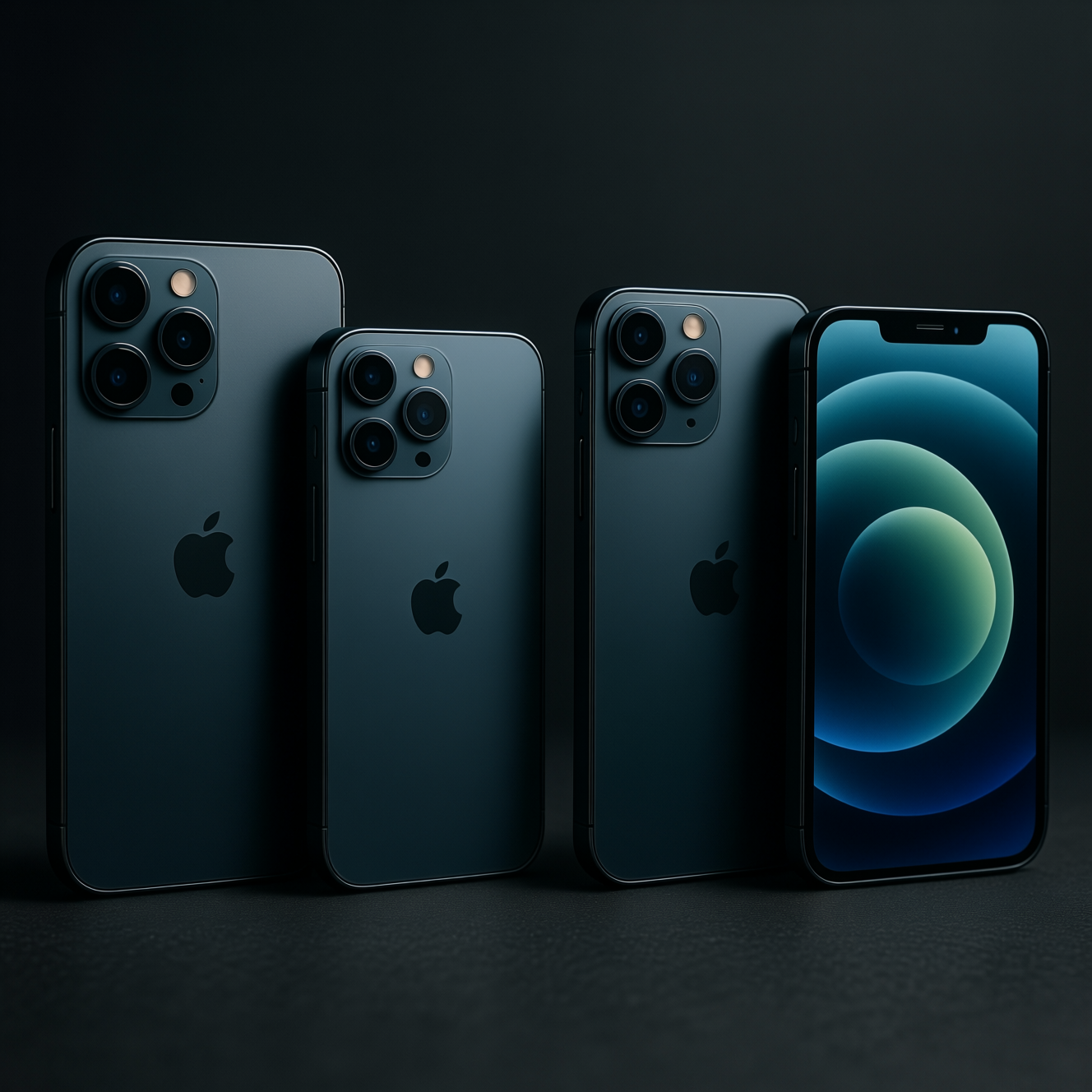}
        \hfill
        \includegraphics[width=0.495\linewidth]{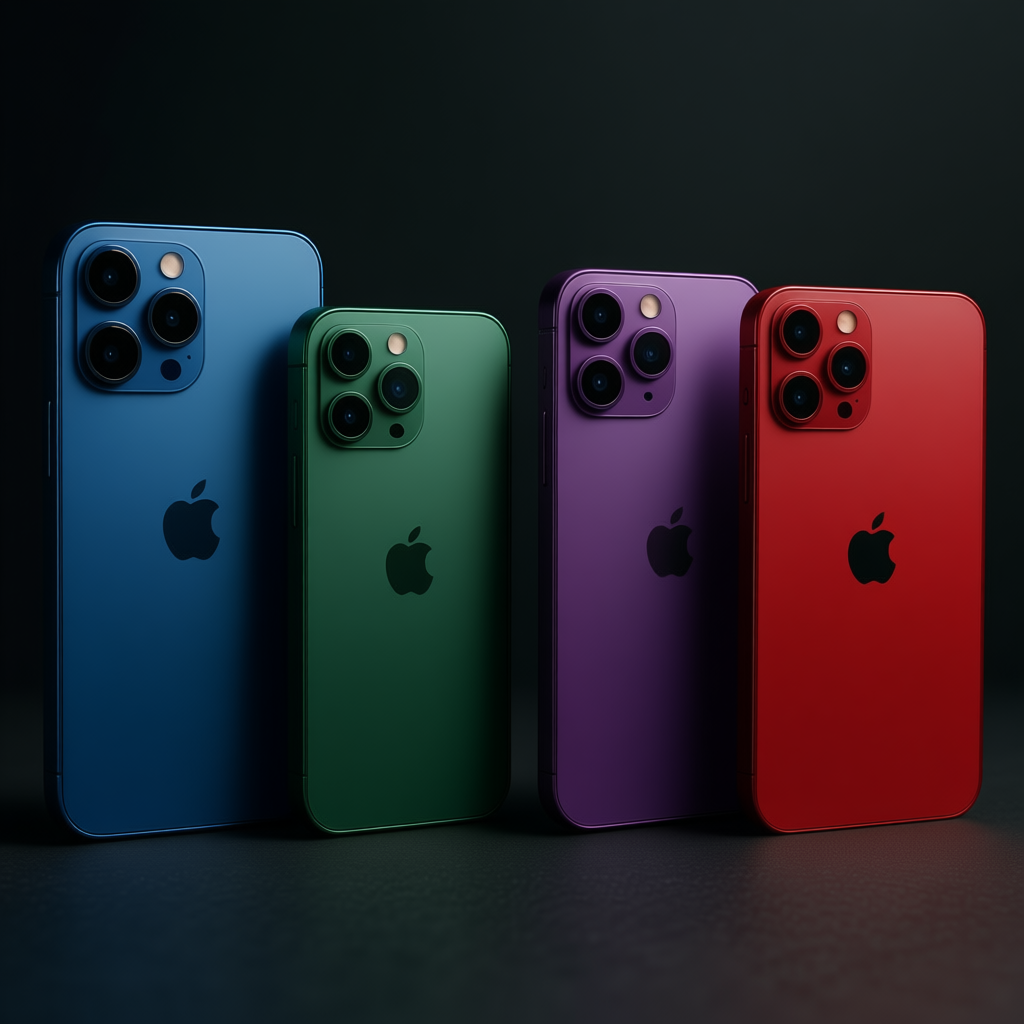}
        \caption{\textbf{Aesthetic Awareness:} M3 infers a need for color diversity to improve aesthetics.}
        \label{fig:phones}
    \end{subfigure}
    
    \vspace{1em}
    
    \begin{subfigure}{\linewidth}
        \centering
        \includegraphics[width=0.495\linewidth]{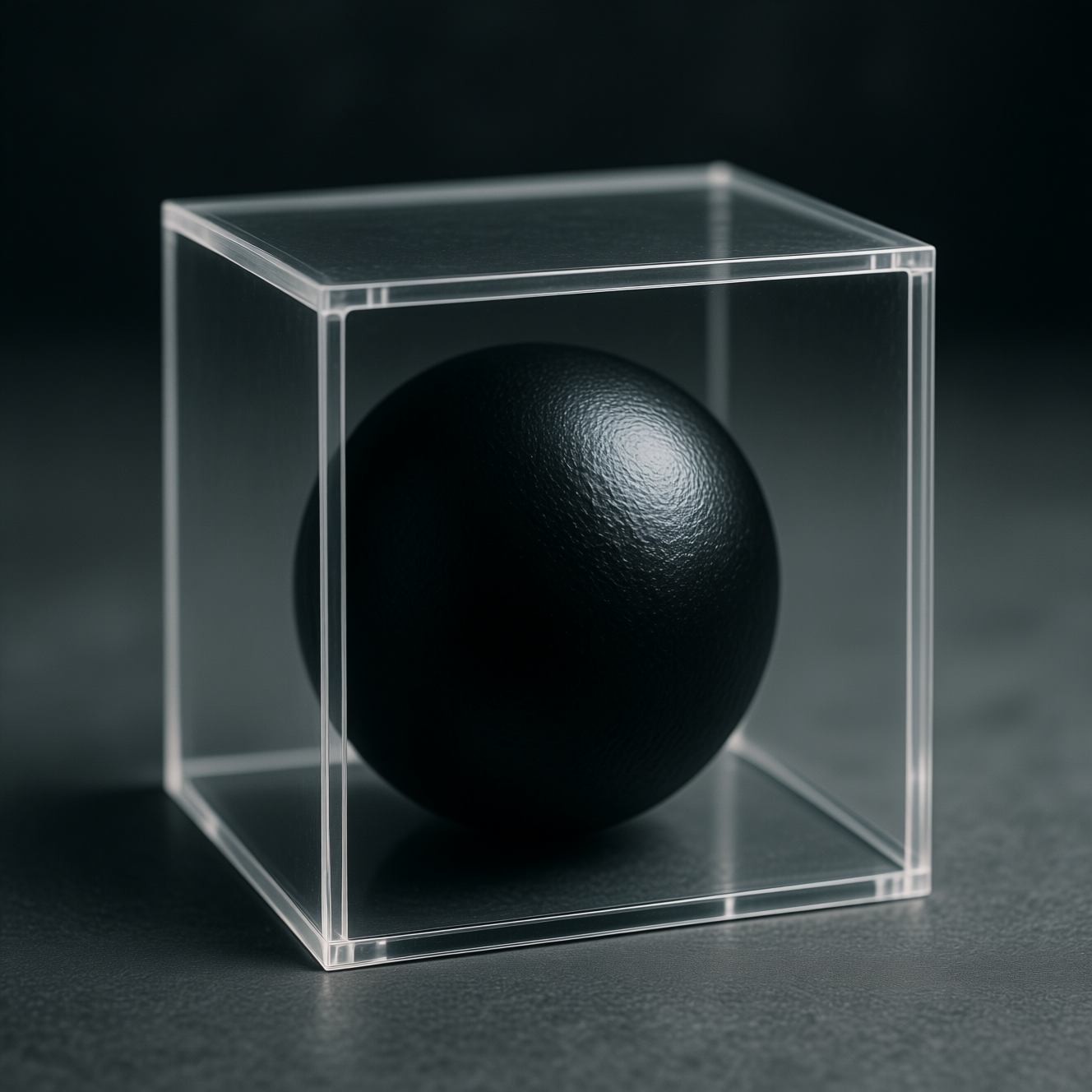}
        \hfill
        \includegraphics[width=0.495\linewidth]{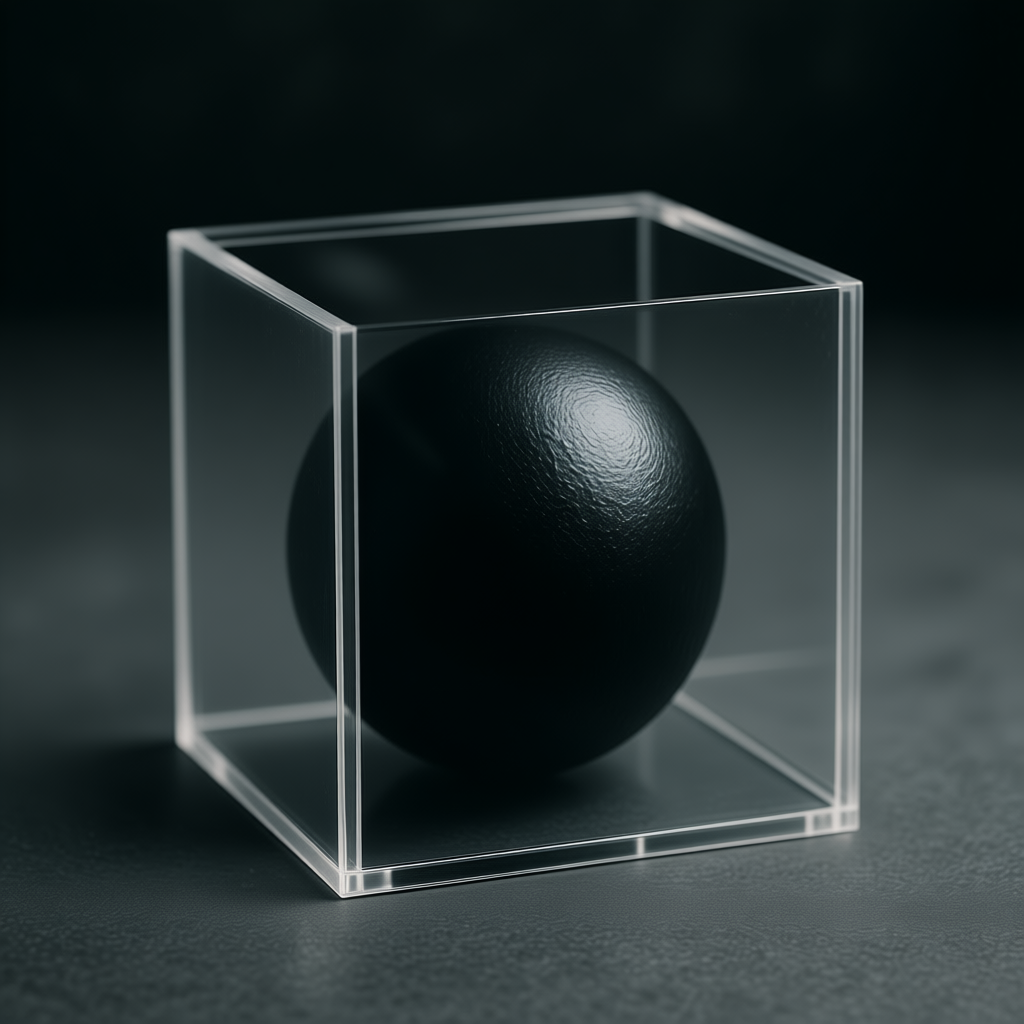}
        \caption{\textbf{Physical Reasoning:} To satisfy ``the ball can easily fall out," M3 reasons that the box must be open.}
        \label{fig:box}
    \end{subfigure}
    \caption{\textbf{Reasoning and Aesthetics.} Examples demonstrating M3's capacity to infer implicit aesthetic details and resolve physical logic.}
    \label{fig:qual_reasoning}
\end{figure}

\noindent\textbf{Case 3: Aesthetic Diversity (Four Cell Phones).}
In the example of ``\textit{a photo of four cell phones}" (Figure \ref{fig:qual_reasoning}(a)), the user did not specify colors. However, M3's Planner inferred an implicit aesthetic goal, noting \textit{``Color diversity (since multiple devices likely have different colors)"} in its checklist. Consequently, it issued the instruction: \textit{``Change the color of the leftmost phone to blue, the second to green, the third to purple, the fourth to red,"} transforming a monotonous image into a vibrant, aesthetically pleasing one.

\noindent\textbf{Case 4: Physical Reasoning (The Ball in a Box).}
Figure \ref{fig:qual_reasoning}(b) illustrates M3's reasoning capabilities. The prompt ``\textit{A black ball is in a transparent box, but the ball can easily fall out}" contains a logical contradiction with the visual of a closed box. M3 recognized that for a ball to ``easily fall out," the container cannot be sealed. It thus generated the physical correction: \textit{``Change the box to have an open top, remove the lid from the transparent container,"} effectively aligning the visual geometry with the text's physical implication.

\clearpage

\section{Complete Case Study of M3-AutoRefiner}
\label{sec:m3_case_study}

\begin{figure*}[p]
    \centering
    \includegraphics[width=\textwidth, height=0.95\textheight, keepaspectratio]{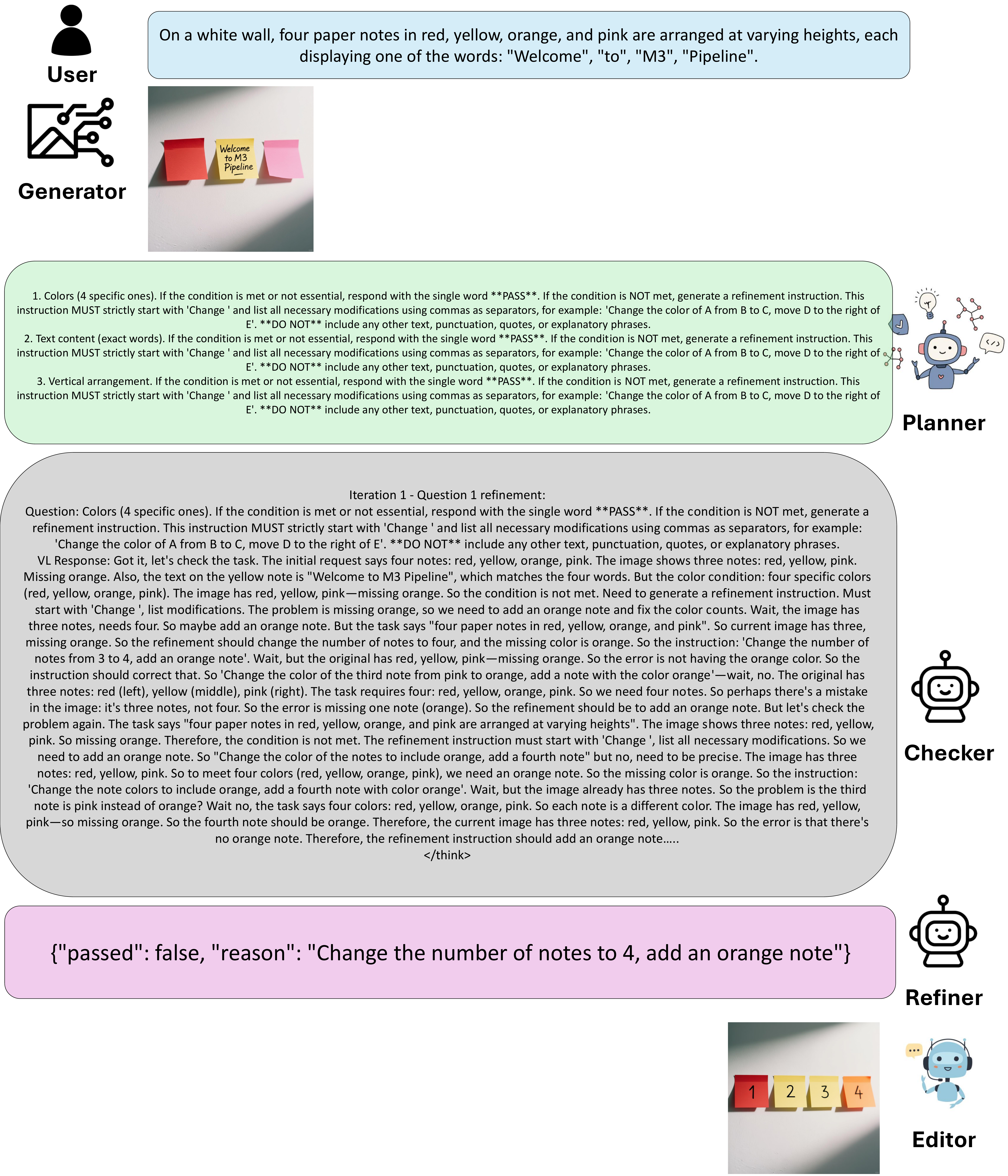}
    \caption{Complete execution trace of M3-AutoRefiner (Part 1/3). This phase demonstrates the initialization and the first round of refinement focused on checking basic constraints.}
    \label{fig:m3_case_part1}
\end{figure*}

\clearpage

\begin{figure*}[p]
    \centering
    \includegraphics[width=\textwidth, height=0.95\textheight, keepaspectratio]{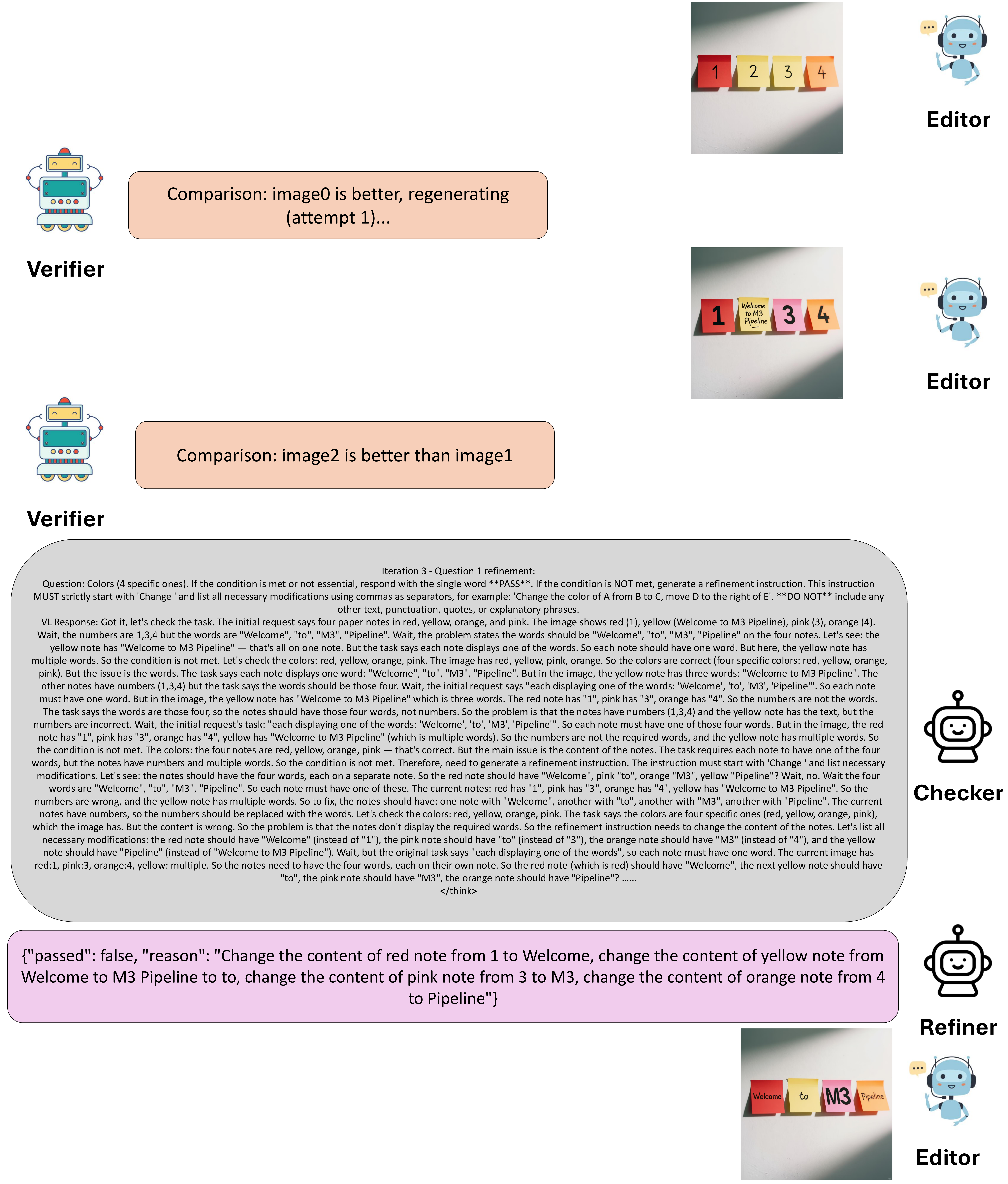}
    \caption{Complete execution trace of M3-AutoRefiner (Part 2/3). The system detects inconsistencies in text content and specific color mappings, triggering the Refiner module.}
    \label{fig:m3_case_part2}
\end{figure*}

\clearpage

\begin{figure*}[p]
    \centering
    \includegraphics[width=\textwidth, height=0.95\textheight, keepaspectratio]{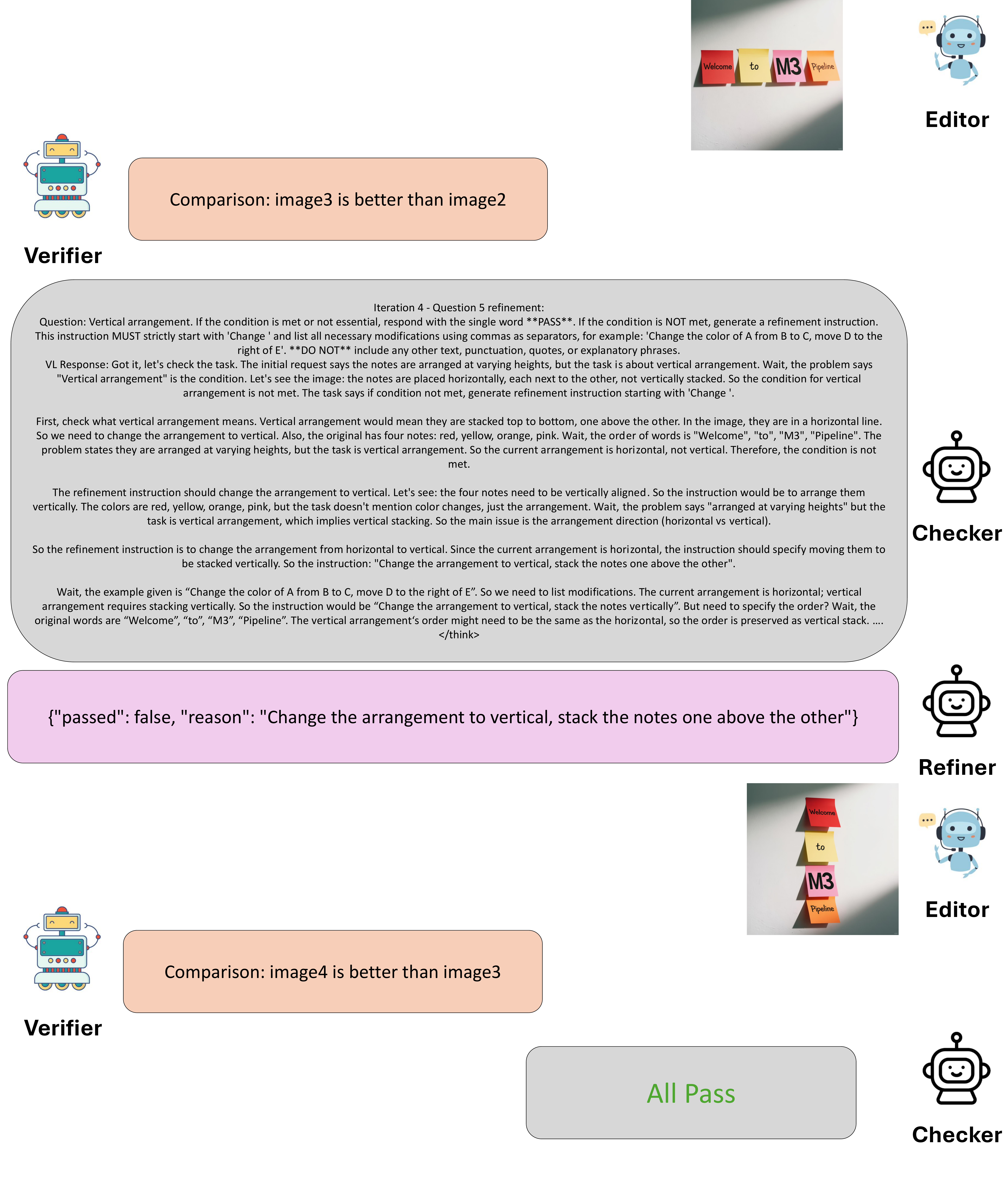}
    \caption{Complete execution trace of M3-AutoRefiner (Part 3/3). The final iteration addresses the vertical arrangement constraint, resulting in the passed verification.}
    \label{fig:m3_case_part3}
\end{figure*}

\clearpage

\end{document}